\documentclass{article} 
\usepackage{times}

\usepackage{hyperref}
\usepackage{url}
\usepackage[pdftex]{graphicx}
\usepackage{tikz}
\usepackage{enumitem}
\usepackage{wrapfig}
\usepackage{natbib}
\usepackage[a4paper, total={6in, 9in}]{geometry}

\title{RL agents Implicitly Learning Human Preferences}

\hypersetup{pdfauthor={Nevan Wichers}, pdftitle={Implicit Human Preferences}}

\author{Nevan Wichers\\
Google Research\\
\texttt{wichersn@google.com} \\}

\begin{document}

\maketitle

\begin{abstract}
In the real world, RL agents should be rewarded for fulfilling human preferences. We show that RL agents implicitly learn the preferences of humans in their environment. Training a classifier to predict if a simulated human's preferences are fulfilled based on the activations of a RL agent's neural network gets .93 AUC. Training a classifier on the raw environment state gets only .8 AUC. Training the classifier off of the RL agent's activations also does much better than training off of activations from an autoencoder. The human preference classifier can be used as the reward function of an RL agent to make RL agent more beneficial for humans.
\end{abstract}

\section{Introduction}
In most RL applications it's assumed that there is a given reward function the agent should optimize. However, in the real world, it is less clear what an agent should be rewarded for, and the wrong choice of reward function could be harmful. We would like RL reward functions to be informed by what humans want. Learning from Human preferences \citep{learning_from_hp} is a possible solution. It trains a neural network to map states of the environment to how much human preferences are satisfied. That neural network is used as a reward function for a RL agent, to train the agent to satisfy human preferences. One problem the authors observed is that the agent exploited inaccuracies in the reward network if the reward network wasn’t updated frequently enough. It will help the learning from human preferences method if the reward function can generalize better based on less data. 

The good news is that a RL agent in the real world will probably have to learn human preferences implicitly in order to achieve its goal, just like how humans have to understand each other's preferences in order to achieve our goals. Even if the RL agent is misaligned with human values, the agent will still have to understand human values in order to determine when the humans will shut it down so it can take precautions. These preferences will be represented by weights in the agent’s neural network. Humans have evolved to represent other people's emotions in their brains \citep{mirror_neurons}, so a RL agent may also.

These human preferences can be extracted from the agent’s neural network to give a mapping between states of the environment and how much human preferences are satisfied. This will probably be better than learning human preferences based on human feedback alone because the RL agent will have seen much more states of the world than can it be labeled by a human. So the agents' understanding of human preferences will likely generalize better. 

An iterative bootstrapping approach can be used to give the agent a reward function aligned with human values. First the agent can be trained with an initial imperfect reward function. Once the agent implicitly learns about human preferences, the reward function can replaced with one based on the extracted human preferences. Then the agent can be trained farther on this new reward function and the process can repeat.

This work is relevant for ML in the real world because
\begin{itemize}[noitemsep,topsep=0pt]
    \item We give evidence that a RL agent in the real world will implicitly learn human preferences, whereas an RL agent in an environment without at least simulated humans won't learn human preferences.
    \item Our method of extracting a reward function based on human preferences would help overcome the challenge of designing a safe reward function in real world RL.
\end{itemize}

We make the following contributions:
\begin{itemize}[noitemsep,topsep=0pt]
  \item We show that RL agents implicitly learn the preferences of humans in their environment. 
  \item We explore different methods to extract the human preferences from the RL agent.
  \item We propose the iterative bootstrapping approach explained above.
\end{itemize}

\section{Method}
We focus on exploring different methods for extracting a model of human preferences from a trained RL agent. \citet{learning_from_hp} already show that if we have a model of human preferences it can be used as a reward function. 

We assume that we have a small amount of supervised data. In our experiments, we assume that the supervised data is pairs of a state of the environment and a boolean representing if human preferences are satisfied or not. However, our approaches can work with other forms of supervised data.

We explore the following techniques to attempt to extract human preferences from the hidden activation of the trained RL model.

\paragraph{Using a single activation}
This technique computes which activation has the highest AUC with the human preferences in the supervised data, and uses that activation as the human preference predictor. This technique will work if a single activation represents how much human preferences are satisfied.

\paragraph{Neural network}
In this technique, a neural network is trained on the supervised data to predict if the human preferences are satisfied given the agent’s hidden activations as input. This will help if human preferences are represented in a more complicated way.

\subsection{Unsupervised}
Since there is only a small amount of labeled data available relative to the number of states the RL agent has visited, unsupervised learning techniques may be helpful.

\paragraph{Clustering}
In this technique, a clustering algorithm is applied to the hidden activations. Then each of the clusters are labeled based on a majority vote of the items in the cluster. Note that this technique will only work if the human preference data is discrete instead of continuous. \citet{transfer_clustering} show that clustering can help with transfer learning.

\paragraph{Dimensionality reduction}
This involves applying dimensionality reduction methods such as PCA or NMF to the hidden activations of the RL agent. After the dimensionality is reduced, any of the techniques described above can be applied on the data with reduced dimensions. \citet{rubik_cube_nmf} had success applying NMF to a neural network trained to solve a Rubik's cube.

\section{Related work}
\paragraph{Multi agent RL}
\citet{agent_model_aux_task} Show that training a RL agent with the auxiliary task of modeling another agent improves performance in multi agent environments. Our work shows that the agent learns to model others even if it isn't trained to.

\citet{Modeling_others_using_oneself} show that if the other agent is sufficiently similar, the agent itself can be used to model the other agent. Our work doesn't make the assumption that the human is similar to it.

\paragraph{RL interpretability} Our work seeks to interpret the hidden activations of a RL agent. There are also other techniques that have been explored for interpreting RL agents.
\citet{vis_atari} and \citet{graying_the_box} apply interpretability techniques like saliency maps, and t-SNE to RL agents. Our work applies interpretability techniques, among others, and does so with the goal of extracting human preferences from the agent.

\paragraph{Theory of mind}
\citet{Machine_Theory_of_Mind} trains a supervised learning algorithm to predict the actions and goals of a RL agent. Our work can be viewed as showing that a RL agent learns an implicit theory of mind without being trained to.

\section{Environment}

\begin{wrapfigure}{r}{0.35\linewidth}
\begin{center}
\includegraphics[width=\linewidth]{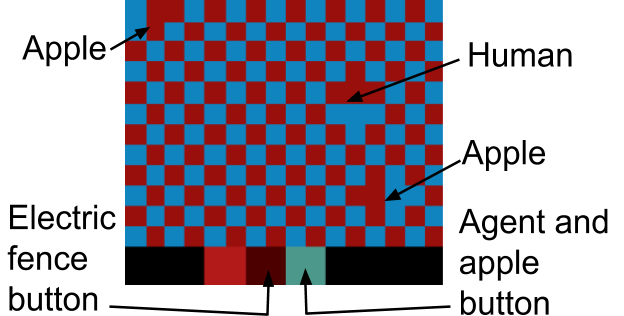}
\caption{Example of the environment}
\end{center}
\label{fig:env_state}
\end{wrapfigure}

We use a grid world environment with a simulated human to compare different methods. The agent is rewarded for pressing a button to collect apples. If the agent has taken too many apples, the simulated human becomes angry and starts taking them from the agent, so the agent loses reward. The way for the agent to prevent this is to press another button to activate an electric fence. The electric fence costs reward at every step, and scares the human away if it tries to take an apple while the fence is active. The fence automatically deactivated after this.

The environment is designed so the optimal strategy for the agent is to only activate the electric fence when it knows the human is about to become angry and start taking apples. If the agent activates the fence too soon, it loses too much reward because of the fence. And if it activates the fence too late, the agent loses reward from the human taking apples. Even though the environment is simple, the fact that the agent has to understand if the human's preferences are satisfied, or if they're angry in order to do well mirrors the real world.

The top part of the environment shows the human and the apples that the agent has collected. The bottom part of the environment shows the agent and also the buttons it can click. The agents actions are to move left or right or to click the button under it. The environment is represented to the RL agent as an image. The human is shown as horizontal stripes and the collected apples as vertical stripes, against a checkerboard background. The colors and the locations of the buttons and apples are randomized to make the environment more diverse and realistic. The episode terminates after a random number of steps, and the time-steps remaining are represented in the state.

\section{Baselines}
\paragraph{Training from the state environment}
This uses the same methods described above only given the environment state as input instead of the activations. We trained a CNN instead of a DNN in this baseline. This will demonstrate if extracting the preferences from the agent instead of the environment state helps. 

\paragraph{Agent not penalized when human takes apples}
The environment is the same except the RL agent isn't penalized when the human takes apples. In this environment, the agent has no reason to care about the simulated human's preferences. If training a human preference model is useful simply because the network extracts useful features, a preference model trained on this agent will work just as well. However, if it's important that the agent has extracted the human preferences instead of only useful features, this won't work as well.

\paragraph{Training from Q values}
This uses the same methods described above only given the Q values of the agent as input instead of the hidden activations of the agent as input.

\paragraph{Autoencoder}
Train an autoencoder to reduce the dimensionality of the image. Then train a network to predict the human preferences from the hidden activations in the middle of the autoencoder.

\section{Experiments}
We trained DQN agents with different hyperparameters and used the network from the agent with the highest reward for these experiments. We then collected a random sample of activations from the last hidden layer of the RL agent to use to train our methods. Each method is trained to predict 0 if the human was scared away by the fence, or is angry, and 1 otherwise.

For each dataset and method we did a random search over hyperparameters using 50 training examples and 100 eval examples. We ran each hyperparameter combination 4 times with different training and eval splits and averaged the results so it wouldn't be too dependent on which set of training examples was chosen. We also used early stopping in each of the techniques to prevent overfitting. The unsupervised methods were trained using about 20,000 unlabeled examples. We trained the model with the best hyperparameters 10 times with 50 training examples and 500 validation examples. For the models trained on activations, we repeated this process 4 times on the 4 best performing RL agents, to make sure that the methods aren't sensitive to noise in the RL training process. An average of the results are reported in table \ref{table:results}.

We tuned the hyperparameters of the autoencoder to reduce the dimensions of the image and decode it with high accuracy. The reconstruction is good enough that one can count the number of apples in the environment. We trained the autoencoder with 4 different random initializations. For each of these we found the best hyperparameters for the neural network trained off of the hidden activations. The results from the autoencoder which got the highest AUC is reported in table \ref{table:results}.

\section{Results}

\begin{table}[h]
\begin{tabular}{c|ccccc}
\tikz{\node[below left, inner sep=1pt] (leftlabel) {Inputs};%
      \node[above right,inner sep=1pt] (toplabel) {Method};%
      \draw (leftlabel.north west|-toplabel.north west) -- (leftlabel.south east-|toplabel.south east);}
                         & NN   & Single & Reduce + NN & Reduce + single \\
Activations (ours)       & 0.93 & 0.88   & 0.92        & 0.87            \\
Image                    & 0.8  & N/A    & 0.54        & 0.5             \\
Q values                 & 0.79 & 0.63   & N/A         & N/A             \\
Activations (no penalty) & 0.79 & 0.73   & 0.77        & 0.76            \\
Image (no penalty)       & 0.78 & N/A    & 0.56        & 0.5             \\
Autoencoder              & 0.6  & N/A    & N/A         & N/A                        
\end{tabular}
\caption{AUC of the experiments. No penalty means the RL agent wasn't penalized when the human ate an apple. Reduce means that a dimensionality reduction techniques was applied before the other method. Single means that the input with the best AUC was used. Activation means the network was trained from the RL agent's activations. Image means the network was trained on the raw state of the environment. N/A means that a certain combination didn't make sense, or wasn't evaluated.}
\label{table:results}
\end{table}

We weren't seeing promising initial results from clustering methods, so we don't show results for them here.

Using the activations got .13 more AUC than using any of the other inputs (.93 vs .8). It's also noteworthy that using only a single activation from the RL agent also got better performance than using any of the other inputs. This means the neural network represents most of the information about human preferences in a single neuron. Dimensionality reduction techniques didn't help the performance. Training from the state of the autoencoder didn't do well.

Training from the image gets about the same result if the agent is penalized for the human taking an apple or not, showing that both of these datasets are the same difficulty.

\section{Conclusion}
Our results suggest that agents implicitly learn about the preferences of humans in their environment, and that extracting those preferences can make predictors more data efficient. We think that this will help agents perform better for humans in the real world, since their reward function will be tied to their robust understanding of human preferences.

In future work we would like to validate our method in more complex environments. We would also like to train an agent using the human preference predictor as the reward function.

\bibliography{iclr2020_conference}
\bibliographystyle{iclr2020_conference}

\end{document}